\documentclass[conference]{IEEEtran}
\IEEEoverridecommandlockouts
\usepackage{cite}
\usepackage{amsmath,amssymb,amsfonts}
\usepackage{algorithm}
\usepackage{algpseudocode}
\usepackage{graphicx}
\usepackage{textcomp}
\usepackage{xcolor}
\usepackage{placeins}
\usepackage{caption}
\usepackage{subcaption}
\usepackage[export]{adjustbox}
\usepackage{wrapfig}
\def\BibTeX{{\rm B\kern-.05em{\sc i\kern-.025em b}\kern-.08em
    T\kern-.1667em\lower.7ex\hbox{E}\kern-.125emX}}
\usepackage{fancyhdr}

\fancypagestyle{firstpage}{
    \fancyhead[L]{}
    \fancyhead[C]{2024 International Conference on Signal Processing and Advance Research in Computing (SPARC)}
    \fancyhead[R]{}
    \fancyfoot[C]{979-8-3503-8520-5/24/\$31.00 ©2024 IEEE}
}
\fancypagestyle{plain}{
    \fancyhead[L]{}
    \fancyhead[C]{}
    \fancyhead[R]{}
    \fancyfoot[C]{}
}

\begin{document}

\title{Enhancing CNN Classification with Lamarckian Memetic Algorithms and Local Search}

\author{\IEEEauthorblockN{Akhilbaran Ghosh}
\IEEEauthorblockA{\textit{Academy Of Technology, Kolkata} \\
\textit{Adisaptagram, Hooghly, India} \\
\textit{akhilbaranghosh@gmail.com}}
\and
\IEEEauthorblockN{Rama Sai Adithya Kalidindi}
\IEEEauthorblockA{\textit{Shiv Nadar University, Dadri} \\
\textit{Uttar Pradesh, India} \\
\textit{kalidindiadithya@gmail.com}}
}

\maketitle
\thispagestyle{firstpage}

\begin{abstract}
Optimization is critical for optimal performance in deep neural networks (DNNs). Traditional gradient-based methods often face challenges like local minima entrapment. This paper explores population-based metaheuristic optimization algorithms for image classification networks. We propose a novel approach integrating a two-stage training technique with population-based optimization algorithms incorporating local search capabilities. Our experiments demonstrate that the proposed method outperforms state-of-the-art gradient-based techniques, such as ADAM, in accuracy and computational efficiency, particularly with high computational complexity and numerous trainable parameters. The results suggest that our approach offers a robust alternative to traditional methods for weight optimization in convolutional neural networks (CNNs). Future work will explore integrating adaptive mechanisms for parameter tuning and applying the proposed method to other types of neural networks and real-time applications.
\end{abstract}


\section{Introduction}
Nature-inspired discoveries and inventions such as solar panels inspired by grove leaves or self-cooling architectures inspired by termite mounds have led scientists to the most obvious decision of seeking to replicate the human brain by building an intelligent system. In 1943, Warren McCulloch and Walter Pitts\cite{fitch_1944}debuted the first Artificial Neural Network design (ANN), also known as Neural Networks (NNs), which used propositional logic to mimic the relationships between neural events and nervous activity. The Multilayer Perceptron (MLP) is the foundational step from ANNs towards Deep Learning (DL). MLP is a feed-forward neural network augmentation in which data is transferred from the input layer to the hidden layers and eventually to the output layer. When presented with fresh, previously unknown data, the MLP can be trained to generalize effectively. As a result, it became a popular choice for creating applications like speech or image recognition and handling complex non-linear problems.\cite{GARDNER} MLPs can generalize well; however, they are not translation invariant, which implies that the system provides the same response no matter how the input is shifted. Convolution Neural Networks (CNNs), a solution to this issue, were introduced by Yann LeCun\cite{Lecun} in 1980 and subsequently improved for handwritten digit recognition between 1988 and 1998. CNNs have been around for a while, but recently, as computer power and data availability have increased, they have emerged as a dominating approach for image recognition and classification. The architecture of CNNs has a significant impact on their power. LeNet-5\cite{Lecun}, which consists of primary convolution, pooling, and fully connected layers, was one of the first CNN designs to be used for handwritten and machine-printed digit recognition.
Deeper CNNs, such as AlexNet\cite{AlexNet} and GoogLeNet\cite{GoogleNet}, have paved the way for computer vision applications. K. Simonyan and A. Zisserman of Oxford University introduced VGGNet\cite{VGG}, a conventional CNN architecture, in two flavors, VGG16 and VGG19. The number following VGG denotes the number of layers used by the architecture. This architecture has lately gained popularity, with over 92\% accuracy in the ImageNet\cite{ImageNet} dataset. The introduction of Residual Network ResNet\cite{ResNet}, a very deep CNN comprised of 152 layers, has solved one of VGGs problems of losing generalization when becoming too deep. The vanishing gradient was one of the VGG bottlenecks because the loss function decreased to minimal values and prevented the weights from altering their values, resulting in no learning. ResNet is famous for solving this problem by using skip connections, which allow the gradients to backpropagate to the initial filters.
\par
Genetic Algorithms (GAs) are heuristic search algorithms influenced by Charles Darwin's theory of natural evolution, which means that new results are created by altering candidate solutions from a population by recombining and mutating them to produce new offspring. This process then repeats for various numbers of generations. Finding a set of inputs to an objective function that yields a maximum or minimal function evaluation is known as optimization. We can use GA to optimize a CNN because training one is an optimization problem. 
Another search technique that draws inspiration from nature and is population-based is Particle Swarm Optimization (PSO)\cite{PSO}, which was initially developed by James Kennedy and Russell Eberhart. The algorithm, which involves particles moving across a search space and updating their velocities by the best-known position, was inspired by the movements of a school of fish or a flock of birds.
Nondominated sorting genetic algorithm NSGA-II\cite{NSGA2}  is a multi-objective optimization algorithm and has been introduced as an improved version of NSGA by Kalyanmoy Deb et al. By integrating the populations of the parents and the offspring, NSGA-II chooses the most Pareto-optimal solutions to address the non-elitism and complexity issues of the original NSGA.
\par
In this paper, we will explore the domain of weight-based optimization of MLP layers of a CNN image classification architecture. We will implement a base CNN architecture pre-trained on the CIFAR10 dataset and then move towards optimizing the weights in the final MLP layer using a population-based memetic algorithm with local search capabilities. This approach is essentially a combination of a pre-trained network and Lamarckian optimization ideology.

\subsection{Literature Review}
This section introduces CNNs and the proposed optimization methods, which include a Gradient Descent (GD) approach, two population-based heuristic algorithms, GA, PSO, and our proposed memetic algorithm.
\subsubsection{Convolutional Neural Networks}
A CNN is composed of basic linked processing units known as neurons, which are modeled after the function of the animal brain. Each neuron gets many inputs, which can also be the outputs of other neurons. Thanks to learnable weights and biases, the weighted dot product of the inputs is then computed and modified using transformation functions such as Sigmoid or Rectified Linear Unit (ReLU). The fundamental CNN design consists of Convolutional, Pooling, and fully connected layers. The Convolution and Pooling layers together with ReLU activation function are mostly utilized to extract features. Edges, points, and different patterns are detected from the original image, which also aids in reducing its dimensionality.  A preset size convolutional filter, also known as a kernel, is applied to the image. This traverses the image's height and width, extracting meaningful data and lowering its dimensionality resulting in a feature map. To further decrease the dimensionality, a Pooling layer is then added. Max-pooling is a popular pooling technique that determines the feature map's maximum value by using a specified receptive field; the most frequent size is 2x2 with stride 2. Finally, the classification of images into various classes is achieved by passing the reduced dimensionality of the image in the form of outputs from previous layers to a multilayer perceptron. A loss function then measures the difference between the expected labeled input instances and the result produced by the learning model to estimate the performance. The model is trained by repeating the preceding steps and modifying the weights to minimize the loss function; this is referred to as training the network. We will next go through several optimization methods that may be used to optimize a convolutional neural network.
\subsubsection{Gradient Descent}
Before the development of modern computers, the first-order iterative optimization algorithm gradient descent was proposed by Augustin-Louis Cauchy in 1847. This approach is currently widely used in Machine Learning (ML) and Deep Learning (DL) to minimize the loss function in several updated forms. Gradient-based approaches' main goal is to determine an optimal location by computing the derivative of an objective function to determine a direction and selecting a step size to denote the leap at each iteration. Depending on the problem, this can be a minimum (gradient descent) or maximum (gradient ascent) location. Vanilla GD, Stochastic GD, and Mini batch GD are three potential GD optimizers that might be employed for our problem.
\subsubsection{Genetic Algorithms}
GA is an iterative method that starts with a randomly produced population of chromosomes and progresses via a process of selection and recombination based on their fitness to produce a new generation. The structure of GA looks as follows. 
\begin{itemize}
    \item Chromosome encoding: a set of candidate solutions to a particular problem.
    \item Fitness: a function to measure the performance of individuals at each iteration.
    \item Selection: a process of selecting parent chromosomes to be used for creating the child population. Selection methods can be Roulette Wheel, Random Stochastic, Tournament and Truncation
    \item Recombination: two main methods of recombination are Crossover and Mutation
    \item Evolution: The above process is reiterated until a stopping criterion is reached
\end{itemize}
    
To reduce computational costs and offer reliable, quick global optimization, PSO does away with the crossover and mutation processes seen in GA algorithms.\cite{loussaief2018convolutional} PSO is frequently used to automate the hyperparameter tuning of CNNs \cite{PSO_CNN}. However, optimizing an objective function poses a difficulty since the location and value of optima might change over time. To achieve good results with PSO, outdated memory and diversity loss problems should be addressed.
\subsubsection{Memetic Algorithm}
Memetic algorithms \cite{Moscato2003} (MA) are similar to GA in that they involve individual learning, social learning, and teaching in addition to mutation recombination and selection. Whereas GA is motivated by biological development, MA is influenced by cultural evolution. To obtain the ideal weights of our convolutional neural network, we used the Lamarckian memetic algorithm in our suggested approach.

\section{CNN Architecture}
The neural network in our experiment was trained using the CIFAR10\cite{CIFAR10} dataset. A total of 60000 color images from 10 distinct classes make up the collection. Each class consists of 6000 pictures that represent various objects, including trucks, vehicles, frogs, horses, deer, cats, birds, and cats. We have randomly selected 40000 images for training the network and 10000 for validation. The remaining 10000 images were used for the final testing.
\par
For the classification of the CIFAR10\cite{CIFAR10} dataset, we employ a basic convolutional neural network composed of three convolution layers and one fully connected layer. To regularise the network, reduce overfitting during the training phase, and expose the model to additional permutations of the same data. The following augmentation processes are employed before feeding the data into the network. To begin, a horizontal flip is performed randomly with a probability of 0.5, inverting the image's rows and columns. The input image is then subjected to a random vertical flip with the same 0.5 chance of flipping the entire image vertically. Finally, with a probability of 0.3, the image is converted to greyscale. Before reaching the network, the image is converted into a three-dimensional tensor with values ranging from -1 to 1. 
The network's initial section, which is utilized for feature extraction, comprises three convolutional layers, each followed by a pooling layer. All convolutional layers have a kernel of the size 3x3 with stride one and padding zero. Batch Normalisation is included in the Conv layers to help stabilize the network during training. Because the range values of the previously extracted features may differ, batch normalization computes the mean and variance of all the extracted features to guarantee that they are all on the same scale. We adopt the rectified linear unit ReLU activation function for computational ease in all convolutional layers. The total number of parameters in the network is 374382.
\begin{figure*}[htp]

\centering\scalebox{0.8}{\includegraphics{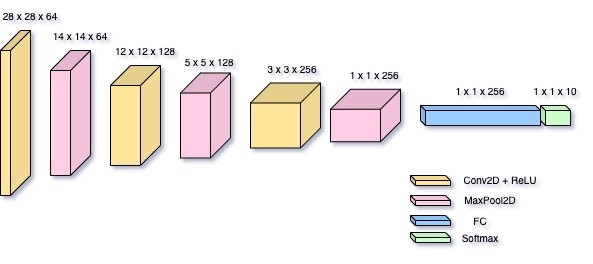}}
\caption{CNN architecture}
\label{fig:cnnj}
\end{figure*}
Because we are using stochastic gradient descent to pre-train our network, ReLU has a significant benefit as it is a nonlinear function that behaves like a linear one, increasing the network's speed. To further minimize the number of parameters and computational effort, the MaxPool2d layer with a 2x2 pooling kernel and zero padding is used. This weightless layer chooses the maximum input value from each receptive area. Further, the results are flattened into a one-dimensional tensor before being fed to the fully connected layer. Finally, the inputs are classified into the predicted classes by using a fully connected linear layer with a softmax activation function. The softmax function converts the one-dimensional tensor into a probability vector containing the probabilities for each of the ten classes. It then returns the highest-valued index from the probability vector.

\section{Training Algorithm}
Our proposed algorithm revolves around the Lamarckian optimization method coupled with local gradient descent optimization along a pre-trained Convolutional Neural Network Architecture. This method falls under the category of memetic algorithms, which belong to a class of stochastic global search techniques that, broadly speaking, combine within the framework of evolutionary algorithms (EAs) the benefits of problem-specific local search heuristics and multi-agent systems\cite{Krasnogor_2012}. For the given problem, the genetic algorithm and PSO lack the local optimization capabilities to find the refined solution for each individual in its exploratory search. Therefore, the use of a more dynamic and robust optimization algorithm is required, which in our case is the  Lamarckian optimization scheme. The following pseudo-code shows the implementation of our chosen algorithm, followed by a brief description of decision justifications of elements within the suggested algorithm.

\begin{algorithm}
\newcommand\tab[1][1cm]{\hspace*{#1}}
\caption{Memetic Lamarckian}\label{alg:cap}
1 Initialise generation 0: \\
\tab g := 0 \\
\tab ${P}_{k}$ := a population of n randomly-generated individuals \\
2 Evaluate the fitness of ${P}_{k}$: \\
\tab Compute fitness(i) \textbf{for} each i {$\in$} ${P}_{k}$ \\
\tab \textbf{While} g {$\leq$} epochs: \\
\tab \tab Selection of parents:\\
\tab \tab \tab {1. Select  fittest parents from $ {P}_{k}$ and \tab \tab \tab { } { } insert   into $ {P}_{k}$+1(offsprings)} \\
\tab \tab Perform crossover(SBX):  \\
\tab \tab \tab 1. Pair them up based on random \tab \tab \tab { }{ }{ } probability \\
\tab \tab \tab 2. Produce offspring \\
\tab \tab \tab 3. Insert the offspring into $ {P}_{k}$+1 \\
\tab \tab Perform mutation(PM): \\ 
\tab \tab \tab 1. Select offsprings in $ {P}_{k}$+1 based \tab \tab \tab {  } { } {} on random probability \\
\tab \tab \tab 2. { }Invert a randomly-selected bit in each \\
\tab \tab Perform Local Search(Lamarkian): \\
\tab \tab \tab 1. Perform local search based on \tab \tab \tab { } {  } {} gradient descent for each  i  {$\in$} $ {P}_{k}$+1 \\
\tab \tab \tab 2. Update the chromosome of each  \tab \tab \tab { } {  } {} i  {$\in$} ${P}_{k}$ +1 \\
\tab \tab Evaluate Fitness each  i  {$\in$} ${P}_{k}$ +1 \\
\tab \tab Survival Selection (offsprings selected for next \tab \tab generation): \\
\tab \tab \tab 1.{} Generational selection \\
\tab \tab Increment to next generation: \\
\tab \tab \tab 1. g += 1
\end{algorithm}

\par
The Lamarckian optimization algorithm is a derivative of evolutionary algorithms combined with a lifetime learning component. The optimization algorithm comprises the following decision choices: Representation, Parent Selection, Genetic Variations, Survival Selection, Local Search Algorithm.

\subsection{Representation:}
The representation/coding is composed of 2 distinct types: Real coded and Binary coded. For our chosen algorithm we will employ the use of real coded representation, the justification for this is that real encoding mitigates the issues caused by fixed string length limits on the precision of the solution as well as the issue of unevenly spaced representation introducing a bias in the search.
\subsection{Parent Selection:}
Roulette Wheel Selection, A variant of the Fitness-proportionate method is used in parent selection along with the concept of elitism. Roulette wheel selection is mainly employed as it gives a chance to all individuals to be selected and thus it preserves diversity while not compromising on the convergence speed unlike tournament selection.\cite{miller1995genetic}. Furthermore, Elitism is also used to ensure that ideal candidates from the parent population are preserved for the next generation to ensure faster convergence based on the fact that the optimization algorithm does not need to waste computational resources to rediscover this solution if it's part of the ideal solution.

\subsection{Genetic Variations:}
    \subsubsection{Cross-over}
    As we are using real-encoded representation, therefore we will be using simulated binary crossover. This method uses a probability density function(beta spread) that simulates the single-point crossover used in binary-coded representation as seen in fig 2.\cite{Deb1995SimulatedBC} 
    \begin{center}
        \begin{figure}{}
        \centering\scalebox{0.4}{\includegraphics{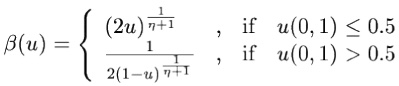}}
        \caption{spread factor}
        \label{fig:wrapfig}
       \end{figure}
     \end{center}
     \vspace{-1em}
    \subsubsection{Mutation}
    We will use the polynomial mutation operator suggested by Deb and Agrawal\cite{deb1999niched} which consists of a user-defined parameter {$\mu M$} ,where {$\mu M$} is in the range [20-100].In addition, p’ is defined as any real-coded decision variable p  [xi (L) , xi (U)] and and is calculated as seen in fig 3 and 4.
        \begin{figure}[!h]
        \centering\scalebox{0.4}{\includegraphics{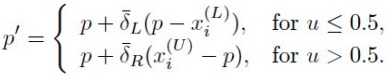}}
        \caption{real coded}
        \label{fig:cnnj}
        \end{figure}
        \FloatBarrier
        \begin{figure}[!h]
        \centering\scalebox{0.4}{\includegraphics{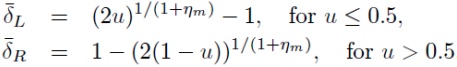}}
        \caption{lower and upper bounds}
        \label{fig:cnnj}
        \end{figure}
        \FloatBarrier
\subsection{Survival Selection}
This component ensures the optimal set of offspring is chosen to be the next set of parent generations. We will use a steady state, choosing the best individual form both the parent and offspring population to subsequently become the next generation's parent population. This approach makes the search space more exploitative and hence helps in converging to an optimal solution in significantly less time complexity in comparison to generational selection methods.
\subsection{Local Search algorithms}
This is the essence of the Lamarckian approach, to achieve a refined solution for each individual. There are several local search algorithms but we will focus solely on gradient-based methods and simplex methods. For our chosen optimization algorithm we have employed gradient-based local search algorithms as the gradient method is recommended for functions with several variables and obtainable partial derivatives\cite{CERDA2016641}. In lieu,  gradient search methods converge at a much faster rate than simplex methods. Furthermore, simplex methods pose the threat of not converging to a solution at all, which is mitigated when using a gradient-based approach.
\begin{figure*}[!ht]

\centering\scalebox{0.8}{\includegraphics{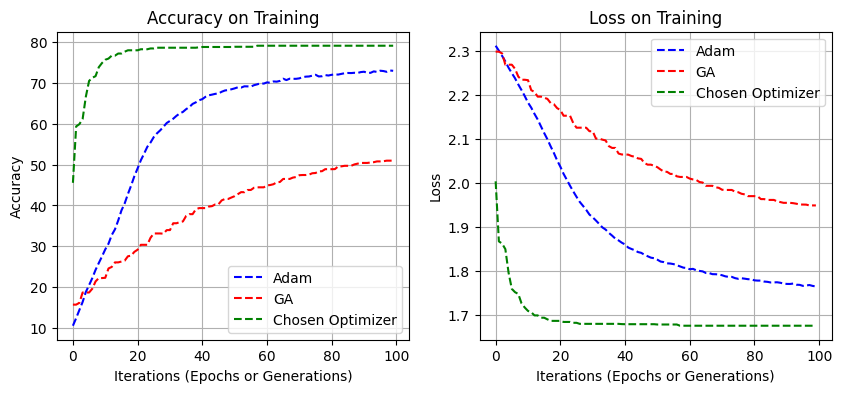}}
\caption {Results}
\label{fig:cnnj}
\end{figure*}
\section{Evaluation}
The initial setup of our experiments comprises a pre-trained convolutional neural network coupled with different weight-based optimization techniques. We have solely focused on 3 types of optimization algorithms: Gradient-based, Genetic and our chosen method. These optimization techniques are implemented on the final MLP layer due to computational complexity constraints. Table I depicts the exact parameters for each optimization experiment having the number of generations (epochs) constant across all techniques. Furthermore, concerning population-based techniques the population size, bounds, elitist constraint, and encoding type are also kept constant. 
\vspace{-4em}

\begin{center}
\begin{table}
\begin{tabular}{ |p{2.5cm}||p{1.5cm}|p{1.5cm}|p{1.5cm}|  }

 \hline
 & Gradient & Populations&Chosen\\
 \hline
 Optimizer   & Adam    &GA&   Memetic\\
 Learning rate&   0.001  &-   &-\\
 Iteration &100 & 100&  100\\
 Population size   &-& 100& 50\\
 Lower bound&   -  & -1 &-1\\
 Upper bound& -  & 1   &1\\
 No of elitists& -  & 1&1\\
 Encoding & -  &  Real Coded& Real Coded\\
 \hline
 \multicolumn{4}{|c|}{Local search} \\
 \hline
 Optimizer& -  & - & Adam\\
 Iteration& -  & -&5\\
 Learning rate& -  & -&0.001\\
 \hline
\end{tabular}
\caption {}
\label{tab1}
\end{table}
\end{center}


\begin{center}

\begin{table}
\begin{tabular}{ |p{2.5cm}||p{1cm}||p{1cm}|p{1cm}| }
\hline

&SDG &GA &Ours\\
\hline
Train Loss&   1.74&1.95&1.67\\
Vall Loss&  1.74&1.95&1.71\\
Train Acc& 75.75&50.97&79.11\\
Vall Acc& 75.18&50.96&74.79\\
\hline
\end{tabular}
\label{tab1}
\caption {}
\end{table}
\end{center}

\par
To perform a comparative analysis on a novel approach for weight-based optimization method coupled with 2-stage training. The results are tabulated in Table II as well as depicted in Figure Y. The initial base method was CNN weight optimization of MLP using a gradient-based approach. Table II shows that the optimal solution reached by the gradient descent method was 75.75 accurate, with a validation accuracy of 75.18. Compared to a genetic algorithm approach, this base method indicates the disadvantages of using genetic optimization to solve this problem. The first issue that presents itself is the convergence time required for the genetic algorithm to reach an optimal solution. Moreover, the optimal solution reached by the genetic algorithm is not comparable to the gradient-based method, as there is a percentage error more significant than 30\%. This furthers the argument that population-based optimization algorithms tend to hinder the computation times and computational accuracy for weight optimization of CNNs due to the sheer number of computations needed to optimize the weights for each individual in the population.
\par
Concerning our chosen algorithm to optimize the weights of the network, we can conclude that our chosen algorithm performs substantially better than its counter-population-based genetic algorithm by an error factor of 35.57. The percentage increase factor, when compared to the percentage difference between the gradient descent method and population-based methods, depicts the competitive advantage of using our chosen algorithm in a time-accuracy trade-off. The difference in performance depends on a few factors, such as the computational complexity posed by this specific problem. To make this problem computationally feasible, we need to employ a method that converges to an optimal solution in a minimal generation. 
\begin{figure*}[!ht]

\centering\scalebox{0.45}{\includegraphics{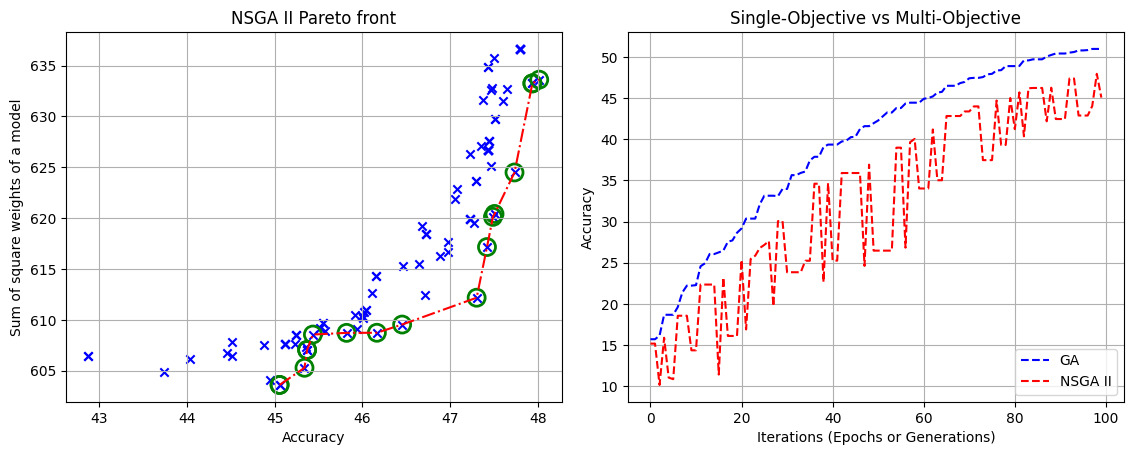}}

\caption{}
\label{fig:cnnj}
\end{figure*}
\par
The memetic algorithm approach we employ has local search capabilities, ensuring an optimal solution for each individual locally. Therefore, the initial accuracy achieved by this method starts at 47\% accuracy at the first generation. Furthermore, in as little as 20 generations, this approach reached the optimal accuracy of 75.75 compared to the gradient descent optimization (ADAM) and genetic algorithm approach, which took 40 and 100 generations, respectively, to reach an optimal solution. In contrast, we can also infer based on the results that for an increased number of generations gradient descent method will outperform our chosen algorithm at a cost of computational time. 

\section{NSGA II}

The notion of applying regularisation terms to a neural network during the training stages ensures that a network generalizes well on unseen data \cite{regularization}. Gaussian Regularisation is inherently treated as another optimization factor, which involves the minimization of the sum square of the weights. To convert this situation into a multi-objective optimization problem, we can employ Fast Non-dominated Sorting Genetic Algorithms (NSGA II) \cite{NSGA2}, which is the successor of the original NSGA \cite{NSGA}, improving on the time complexity issue posed by the original NSGA. The selection process of an optimal Pareto set in NSGA II depends on selecting a solution in the non-dominated Pareto front concerning the maximization of crowding distance within the selected Pareto front.

\begin{center}
\begin{table}[!ht]
\begin{tabular}{ |p{3.5cm}||p{3cm}| }
\hline
Objective &Optimisation\\
\hline
Accuracy of a network&  Maximising\\
Gaussian regulariser& Minimising\\
\hline
\end{tabular}
\label{tab1}
\caption{}
\end{table}
\end{center}
\begin{center}
\vspace{-2em}
\begin{table}[!ht]
\begin{tabular}{ |p{3.5cm}||p{3cm}| }
\hline
Hyperparameters &Values\\
\hline
Iterations&  100\\
Population size& 100\\
Lower bound& -1\\
Upper bound& 1\\
Encoding &Real coded \\
\hline
\end{tabular}
\label{tab1}
\caption{}
\end{table}
\end{center}
\vspace{-2em}

\par
To treat this problem as a multi-objective optimization problem, we will aim to simultaneously maximize the accuracy while minimizing the sum square of weights. During multi-objective optimization, we will perform 100 generations on a population size of 100 individuals. The bound range will be normalized within the range of -1 to 1, coupled with the real encoded representation as seen in Table IV.
\par
The coverage rate of NSGA-II based on multi-objective optimization is comparatively slower than single-objective optimization. This is because it focuses on optimizing two objective functions instead of a single objective function; it needs to find the most optimal solution in the first Pareto front by measuring the spread of the solution in that front. This is known as the crowding distance. The objective of this network is to achieve the highest accuracy in image classification. By having a second objective, the NSGA-II does not always select the optimal solution based on the accuracy objective but instead selects the solution that has the optimal solution based on the first non-dominated front having the maximal crowding distance. This depicts the issue that if the two objective functions are not linearly proportional, one of the objectives will hinder the other objective from achieving an optimal solution.
\par
In our case of optimization based on the maximization of accuracy as a single objective, we can safely conclude, based on Fig 6, that the objective function (accuracy) is more suited in this problem to be dealt with as a single objective optimization problem. When we experimented with multi-objective optimization, we concluded that the convergence rate derived during multi-objective optimization was significantly slower than single-objective optimization. Fig 6 shows that the accuracy fluctuates because of the dual objective selection process in NSGA-II. Furthermore, let us consider convergence based on optimal accuracy achieved. It is safe to assume that the optimization of weights in an image classification problem such as ours will achieve a more optimal solution with single objective optimization. 

\section{Conclusion}
In conclusion, a 2-stage training approach has been implemented to counter the expansivity of searching space by pre-training a Convolutional Neural Network and employing it as a feature extractor. This feature extractor passes features to a Multi-Layer Perceptron (MLP) to classify the labels of images by using population-based optimizers. The experiment shows that population-based metaheuristic optimization algorithms such as Genetic Algorithms (GA) and Particle Swarm Optimization (PSO) were outperformed by Gradient-based optimization in both computational complexity and efficiency. However, population-based metaheuristics, with the help of local searches, such as memetic algorithms, converge significantly faster than gradient-based methods. Individual learning in a type of local search allows the individual to converge to the optimal solution much faster since it converges to an optimal solution for each individual's local search. Therefore, ensuring that each generation has a near-optimal individual. This approach to refining individuals in a population provides an optimal method of optimization of weights for image classification tasks with population-based optimization methods. The results and findings confer our initial assumption that local search capabilities in population-based optimization significantly improve population-based optimization methods. Our chosen optimization techniques outperform state-of-the-art gradient descent methods in the constraint of 100 epochs/generations. In comparison, genetic algorithms lacking local searching capabilities show a 35\% loss in optimal accuracy achieved in 100 generations. 
\par
In addition, we performed an ablation study to further optimize the problem by checking if this specific image classification problem is a single-objective or a multi-objective optimization problem. In doing so, we concluded that this problem is suited for single objective optimization for achieving an optimal solution in minimal convergence time. The addition of regularisation to the objective function might hinder the convergence of the network, but this is only sometimes the case. Regularisation techniques are applied to significantly improve the network in other domains, such as regularisation in network architectures and data regularisation techniques in the preprocessing stages. 
\par
\subsection{Future Directions }
Due to the time constraint, the research employs a pre-trained CNN feature extractor trained on CIFAR10 dataset. This introduces a bias due to the network being pre-trained on the targeted dataset. Therefore, we propose that future research can revolve around transfer learning coupled with different population-based optimization techniques instead of using a pre-trained feature extractor. Likewise, the same base principles can be applied to look into unsupervised auto-encoder as a feature extractor instead of a pre-trained image classification network. Moreover, the domain of meta-heuristic population-based algorithms in conjunction with hybrid CNN models can be explored to achieve an understanding of multilayer weight optimization with hybrid neural networks. Finally, with the advent of transformers in computer vision-related tasks, combining vision transformers with evolutionary algorithm techniques is currently an unexplored domain.

\bibliographystyle{IEEEtran}
\bibliography{references}

\begin{thebibliography}{10}
\providecommand{\url}[1]{#1}
\csname url@samestyle\endcsname
\providecommand{\newblock}{\relax}
\providecommand{\bibinfo}[2]{#2}
\providecommand{\BIBentrySTDinterwordspacing}{\spaceskip=0pt\relax}
\providecommand{\BIBentryALTinterwordstretchfactor}{4}
\providecommand{\BIBentryALTinterwordspacing}{\spaceskip=\fontdimen2\font plus
\BIBentryALTinterwordstretchfactor\fontdimen3\font minus
  \fontdimen4\font\relax}
\providecommand{\BIBforeignlanguage}[2]{{%
\expandafter\ifx\csname l@#1\endcsname\relax
\typeout{** WARNING: IEEEtran.bst: No hyphenation pattern has been}%
\typeout{** loaded for the language `#1'. Using the pattern for}%
\typeout{** the default language instead.}%
\else
\language=\csname l@#1\endcsname
\fi
#2}}
\providecommand{\BIBdecl}{\relax}
\BIBdecl

\bibitem{fitch_1944}
F.~B. Fitch, ``Warren s. mcculloch and walter pitts. a logical calculus of the
  ideas immanent in nervous activity. bulletin of mathematical biophysics, vol.
  5 (1943), pp. 115–133.'' \emph{Journal of Symbolic Logic}, vol.~9, no.~2,
  p. 49–50, 1944.

\bibitem{GARDNER}
\BIBentryALTinterwordspacing
M.~Gardner and S.~Dorling, ``Artificial neural networks (the multilayer
  perceptron)—a review of applications in the atmospheric sciences,''
  \emph{Atmospheric Environment}, vol.~32, no.~14, pp. 2627--2636, 1998.
  [Online]. Available:
  \url{https://www.sciencedirect.com/science/article/pii/S1352231097004470}
\BIBentrySTDinterwordspacing

\bibitem{Lecun}
Y.~Lecun, L.~Bottou, Y.~Bengio, and P.~Haffner, ``Gradient-based learning
  applied to document recognition,'' \emph{Proceedings of the IEEE}, vol.~86,
  no.~11, pp. 2278--2324, 1998.

\bibitem{AlexNet}
A.~Krizhevsky, I.~Sutskever, and G.~E. Hinton, ``Imagenet classification with
  deep convolutional neural networks,'' in \emph{Advances in Neural Information
  Processing Systems}, F.~Pereira, C.~Burges, L.~Bottou, and K.~Weinberger,
  Eds., vol.~25.\hskip 1em plus 0.5em minus 0.4em\relax Curran Associates,
  Inc., 2012.

\bibitem{GoogleNet}
\BIBentryALTinterwordspacing
C.~Szegedy, W.~Liu, Y.~Jia, P.~Sermanet, S.~Reed, D.~Anguelov, D.~Erhan,
  V.~Vanhoucke, and A.~Rabinovich, ``Going deeper with convolutions,'' 2014.
  [Online]. Available: \url{https://arxiv.org/abs/1409.4842}
\BIBentrySTDinterwordspacing

\bibitem{VGG}
\BIBentryALTinterwordspacing
K.~Simonyan and A.~Zisserman, ``Very deep convolutional networks for
  large-scale image recognition,'' 2014. [Online]. Available:
  \url{https://arxiv.org/abs/1409.1556}
\BIBentrySTDinterwordspacing

\bibitem{ImageNet}
J.~Deng, W.~Dong, R.~Socher, L.-J. Li, K.~Li, and L.~Fei-Fei, ``Imagenet: A
  large-scale hierarchical image database,'' in \emph{2009 IEEE Conference on
  Computer Vision and Pattern Recognition}, 2009, pp. 248--255.

\bibitem{ResNet}
\BIBentryALTinterwordspacing
K.~He, X.~Zhang, S.~Ren, and J.~Sun, ``Deep residual learning for image
  recognition,'' 2015. [Online]. Available:
  \url{https://arxiv.org/abs/1512.03385}
\BIBentrySTDinterwordspacing

\bibitem{PSO}
J.~Kennedy and R.~Eberhart, ``Particle swarm optimization,'' in
  \emph{Proceedings of ICNN'95 - International Conference on Neural Networks},
  vol.~4, 1995, pp. 1942--1948 vol.4.

\bibitem{NSGA2}
K.~Deb, A.~Pratap, S.~Agarwal, and T.~Meyarivan, ``A fast and elitist
  multiobjective genetic algorithm: Nsga-ii,'' \emph{IEEE Transactions on
  Evolutionary Computation}, vol.~6, no.~2, pp. 182--197, 2002.

\bibitem{loussaief2018convolutional}
S.~Loussaief and A.~Abdelkrim, ``Convolutional neural network hyper-parameters
  optimization based on genetic algorithms,'' \emph{International Journal of
  Advanced Computer Science and Applications}, vol.~9, no.~10, 2018.

\bibitem{PSO_CNN}
T.-Y. Kim and S.-B. Cho, ``Particle swarm optimization-based cnn-lstm networks
  for forecasting energy consumption,'' in \emph{2019 IEEE Congress on
  Evolutionary Computation (CEC)}, 2019, pp. 1510--1516.

\bibitem{Moscato2003}
P.~Moscato and C.~Cotta, \emph{A Gentle Introduction to Memetic
  Algorithms}.\hskip 1em plus 0.5em minus 0.4em\relax Boston, MA: Springer US,
  2003, pp. 105--144.

\bibitem{CIFAR10}
\BIBentryALTinterwordspacing
A.~Krizhevsky, V.~Nair, and G.~Hinton, ``Cifar-10 (canadian institute for
  advanced research).'' [Online]. Available:
  \url{http://www.cs.toronto.edu/~kriz/cifar.html}
\BIBentrySTDinterwordspacing

\bibitem{Krasnogor_2012}
N.~Krasnogor, ``Memetic algorithms,'' in \emph{Handbook of Natural
  Computing}.\hskip 1em plus 0.5em minus 0.4em\relax Springer Berlin
  Heidelberg, 2012, pp. 905--935.

\bibitem{miller1995genetic}
B.~L. Miller, D.~E. Goldberg \emph{et~al.}, ``Genetic algorithms, tournament
  selection, and the effects of noise,'' \emph{Complex systems}, vol.~9, no.~3,
  pp. 193--212, 1995.

\bibitem{Deb1995SimulatedBC}
K.~Deb and R.~B. Agrawal, ``Simulated binary crossover for continuous search
  space,'' \emph{Complex Syst.}, vol.~9, 1995.

\bibitem{deb1999niched}
K.~Deb and S.~Agrawal, ``A niched-penalty approach for constraint handling in
  genetic algorithms,'' in \emph{Artificial neural nets and genetic
  algorithms}.\hskip 1em plus 0.5em minus 0.4em\relax Springer, 1999, pp.
  235--243.

\bibitem{CERDA2016641}
\BIBentryALTinterwordspacing
V.~Cerdà, J.~L. Cerdà, and A.~M. Idris, ``Optimization using the gradient and
  simplex methods,'' \emph{Talanta}, vol. 148, pp. 641--648, 2016. [Online].
  Available:
  \url{https://www.sciencedirect.com/science/article/pii/S0039914015300175}
\BIBentrySTDinterwordspacing

\bibitem{regularization}
\BIBentryALTinterwordspacing
J.~Kukačka, V.~Golkov, and D.~Cremers, ``Regularization for deep learning: A
  taxonomy,'' 2017. [Online]. Available: \url{https://arxiv.org/abs/1710.10686}
\BIBentrySTDinterwordspacing

\bibitem{NSGA}
N.~Srinivas and K.~Deb, ``Multi-objective function optimization using
  non-dominated sorting genetic algorithms,'' vol.~2, 11 1994.

\end{thebibliography}
\vspace{2em}
\appendix

A
\\
\begin{figure}[!ht]

\centering\scalebox{0.8}{\includegraphics{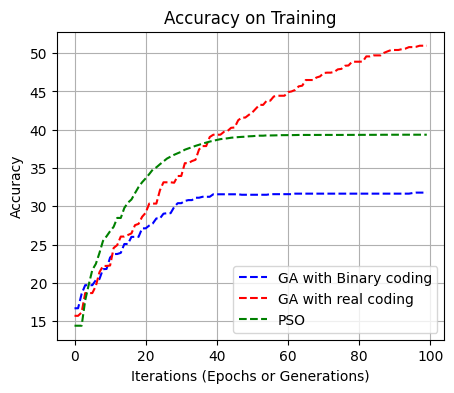}}

\caption{}
\label{fig:cnnj}
\end{figure}
\par
\noindent
Fig.7 compares the results of two population-based optimizers, PSO and GA, with two different encodings for GA. 

\vspace{2em}

B
\begin{figure}[!ht]

\centering\scalebox{0.8}{\includegraphics{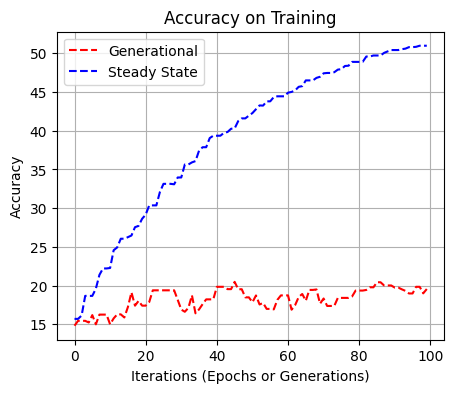}}

\caption{}
\label{fig:cnnj}
\end{figure}
\par
\noindent
Fig. 8 compares the Generational and steady state in the survival selection of genetic algorithms. The steady state converges to an optimal solution much faster than generational.

C
\\
\vspace{-1em}
\begin{figure}[!ht]

\centering\scalebox{0.32}{\includegraphics{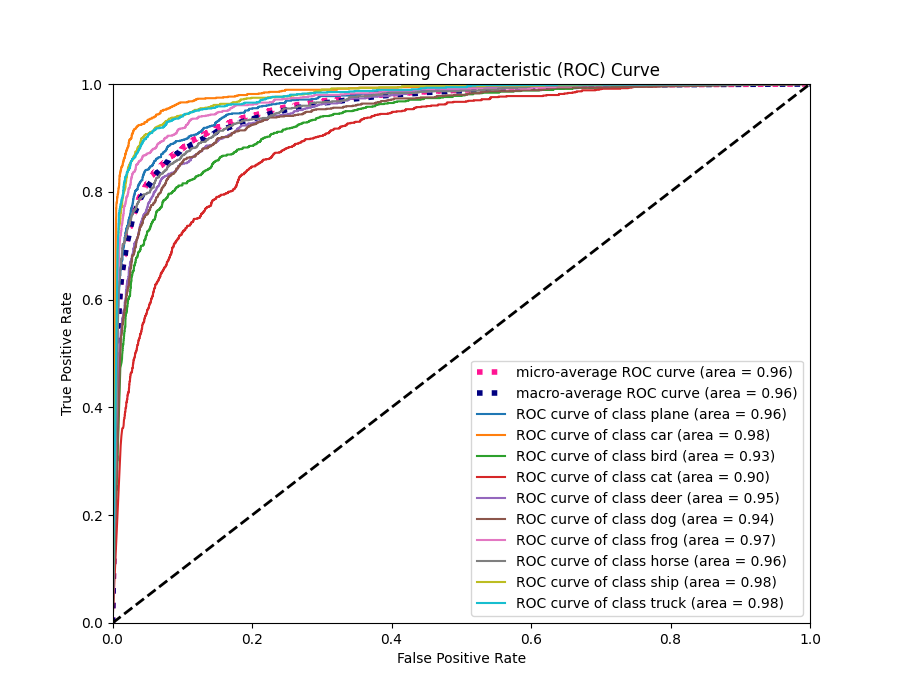}}

\caption{}
\label{fig:cnnj}
\end{figure}
\par
\noindent
Fig. 9 above contains the ROC curve of the classifier that was trained by using memetic algorithms. It shows the area under the curve of the prediction of each class to reflect the performance of a model to distinguish between each class.\\ 
\par
D
\begin{figure}[!ht]

\centering\scalebox{0.8}{\includegraphics{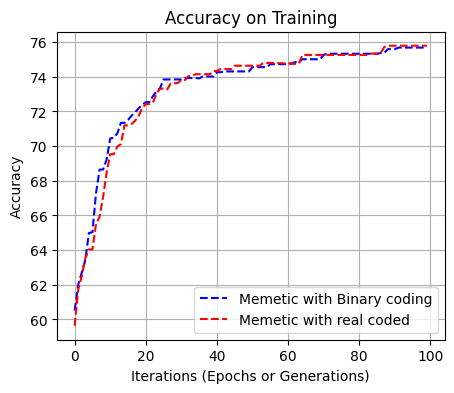}}

\caption{}
\label{fig:cnnj}
\end{figure}
\par
\noindent
Fig. 10 shows the performance of two memetic algorithms with binary (grey) encoding and real-coded encoding. It demonstrates that representation plays an insignificant role in optimisation for memetic algorithms.
\end{document}